\documentclass[11pt]{article}
\usepackage{acl2016}
\usepackage{times}
\usepackage[hyphens]{url}
\usepackage{latexsym}
\usepackage{mathtools}

\aclfinalcopy 
\pdfoutput=1

\title{Translation Quality Estimation using Recurrent Neural Network}

\author{Raj Nath Patel \\
  CDAC Mumbai, India \\
  {\tt rajnathp@cdac.in} \\\And
  Sasikumar M \\
  CDAC Mumbai, India \\
  {\tt sasi@cdac.in} \\}
\date{}

\begin{document}
\maketitle
\begin{abstract}
This paper describes our submission to the shared task on word/phrase level Quality Estimation (QE) in the First Conference on Statistical Machine Translation (WMT16). The objective of the shared task was to predict if the given word/phrase is a correct/incorrect (OK/BAD) translation in the given sentence. In this paper, we propose a novel approach for word level Quality Estimation using Recurrent Neural Network Language Model (RNN-LM) architecture.
RNN-LMs have been found very effective in different Natural Language Processing (NLP) applications. RNN-LM is mainly used for vector space language modeling for different NLP problems. For this task, we modify the architecture of RNN-LM.
The modified system predicts a label (OK/BAD) in the slot rather than predicting the word. The input to the system is a word sequence, similar to the standard RNN-LM. The approach is language independent and requires only the translated text for QE. To estimate the phrase level quality, we use the output of the word level QE system.
\end{abstract}


\section{Introduction}
\label{sec:intro}
Quality estimation is the process to predict the quality of translation without any reference translation~\cite{blatz:2004,specia:2009}. Whereas, Machine Translation (MT) system evaluation does require references (human translation). QE could be done at word, phrase, sentence or document level. This paper describes the submission to the shared task on word and phrase level QE (Task 2) for English-German (en-de) MT. The shared task has the trace of last five years' research in the field of QE~\cite{Callison:2012,Bojar:2013,Bojar:2014,Bojar:2015}.

In recent years, RNN-LM has demonstrated exceptional performance in a variety of NLP applications~\cite{Mikolov:RNNLM:2010,Mikolov:MT:2013,Mikolov:MT1:2013,Socher:Parsing:2013,Socher:Sentiment:2013}.
The RNN-LM represents each word as high-dimensional real-valued vectors, like the other continuous space language models such as feed forward neural network language models~\cite{Schwenk:2002,Bengio:2003,Morin:2005,Schwenk:2007} and Hierarchical Log-Bi-linear language models~\cite{Mnih:2009}.

In this paper, we have used a modified version of RNN-LM, which accepts the word sequence (context window) as input and predicts label at the output for the middle word. For example, let us consider the following input/output sample: 

English (MT input): Layer effects are retained by default .

German (MT output): " Effekte sind standardmäßig beibehalten .

German (Post-edited): Ebeneneffekte werden standardmäßig beibehalten .

Tags: BAD BAD BAD OK OK OK

Now if we have to predict the output tag (BAD) for the word ``sind'' in the MT output, our input sequence to the RNN-LM will be ``Effekte sind standardmäßig'' (if context window size is 3). Whereas, for standard RNN-LM model, ``Effekte standardmäßig'' would be the input to the network with ``sind'' as the output.
We add padding at the start and end of the sentence according to the context window. The detailed description of the model and its implementation is given in section 3.

We have used the data provided by the organizers for the shared task on quality estimation (2016) which includes: 
(i) source sentence
(ii) translated output (word/phrase level) 
(iii) word/phrase level tagging (OK/BAD) 
(iv) post edited translation 
(v) 22 baseline features
(vi) word alignment.
The goal of the task is to predict whether the given word/phrase is a correct/incorrect (OK/BAD) translation in the given sentence. 

The remainder of the paper is organised as follows. Section~\ref{sec:bg} describes the related work. Section~\ref{sec:models} presents RNN models we use, and its implementation. 
In section~\ref{sec:results}, we discuss the data distribution, our approaches, and results. Discussion of our methodology and different models is covered in section~\ref{sec:discussion} followed by concluding remarks in section~\ref{sec:conc}.

\section{Related Work}
\label{sec:bg}
For word level QE, supervised classification techniques are being used widely. Most of these approaches require manually designed features~\cite{Bojar:2014}, similar to the feature set provided by the organizers.

\newcite{logacheva:2015} modeled the word level QE using the CRF++ tool with data selection and data bootstrapping in which data selection filters out the sentences having the smallest proportion of erroneous tokens and are assumed to be less useful for the task. The bootstrapping technique creates additional data instances and boosts the importance of BAD labels occurring in the training data. \newcite{shang:2015} tried to solve the problem of label imbalance with creating sub-labels like OK\_B (begin), OK\_I (intermediate), OK\_E (end). \newcite{shah:2015} have used word embedding as an additional feature (+25 features) with SVM classifier. Bilingual Deep Neural Network (DNN) based model for word level QE was proposed by \newcite{kreutzer:2015}, in which word embedding was pre-trained and fine-tuned with other parameters of the network using stochastic gradient descent. \newcite{de:2014} have used Bidirectional LSTM as a classifier for word level QE.

The architecture of RNN-LM has been used for Natural Language Understanding (NLU) ~\cite{Yao:2013,Yao:2014} earlier. Our approach is quite similar to the~\newcite{kreutzer:2015}, but we are using RNN instead of DNN. We have also tried to address the problem of label-imbalance, introducing sub-labels as suggested by Shang et al. (2015).

\section{RNN Models for QE}
\label{sec:models}
For this task, we exploited RNN's extensions, Long Short-Term Memory (LSTM)~\cite{Hochreiter:1997} and Gated Recurrent Unit (GRU)~\cite{Cho:2014}. LSTM and GRU have shown to perform better at modeling the long-range dependencies in the data than the simple RNN. Simple RNN also suffers from the problem of exploding and vanishing gradient~\cite{Bengio:1994}. LSTM and GRU tackle this problem by introducing a gating mechanism. LSTM includes input, output and forget gates with a memory cell, whereas GRU has reset and update gates only (no memory cell). The detailed description of each model is given in the following subsections.

\subsection{LSTM}
Different researchers use slightly different LSTM variants~\cite{Graves:2013,Yao:2014,Jozefowicz:2015}.
We implemented the version of LSTM described by the following set of equations:
\begin{flalign*}
& i_t = sigm(W_{xi}x_t + W_{hi}h_{t-1} + b_i) \\
& o_t = sigm(W_{xo}x_t + W_{ho}h_{t-1} + b_o) \\
& f_t = sigm(W_{xf}x_t + W_{hf}h_{t-1} + b_f) \\
& j_t = tanh(W_{xj}x_t + W_{hj}h_{t-1} + b_j) \\
& c_t = c_{t-1} \odot f_t + i_t \odot j_t \\
& h_t = tanh(c_t) \odot o_t 
\end{flalign*}
where $sigm$ is the logistic sigmoid function and $tanh$ is the hyperbolic tangent function to add non linearity in the network. 
$\odot$ is the element-wise multiplication of vectors.
$i$, $o$, $f$ are $input$, $output$, $forget$ gates respectively, $j$ is the new memory content whereas $c$ is the updated memory content.
In these equations, $W_*$ are the weight matrices and $b_*$ are the bias vectors.

\subsection{Deep LSTM}
In this paper, we have used deep LSTM with two layers. Deep LSTM is created by stacking multiple LSTMs on the top of each other. 
We feed the output of the lower LSTM as the input to the upper LSTM. For example, if $h_t$ is the output of the lower LSTM, we apply a matrix transform to form the input $x_t$ for the upper LSTM. The matrix transformation allows having two consecutive LSTM layers of different sizes.

\subsection{GRU}
GRU is an architecture, which is quite similar to the LSTM. \newcite{Chung:2014} found that GRU outperforms LSTM on a suit of tasks. GRU is defined by the following set of equations:
\begin{flalign*}
& r_t = sigm(W_{xr}x_t + W_{hr}h_{t-1} + b_r) \\
& z_t = sigm(W_{xz}x_t + W_{hz}h_{t-1} + b_z) \\
& \widetilde{h}_t = tanh(W_{xh}x_t + W_{hh}(r_t \odot h_{t-1}) + b_h) \\
& h_t = z_t \odot h_{t-1} + (1 - z_t) \odot \widetilde{h}_t
\end{flalign*}
In the above equations, $W_*$ are the weight matrices and $b_*$ are the bias vectors. 
$r$ and $z$ are known as the reset and update gate respectively. GRU does not use any separate memory cell as used in LSTM.
However, gated mechanism controls the flow of information in the unit.

\subsection{Implementation Details}
We implemented all the models (LSTM, deep LSTM and GRU) with \footnotemark THEANO \footnotetext{\url{http://deeplearning.net/software/theano/\#download}} framework~\cite{Bergstra:2010,Bastien:2012} as described above. For all the models in the paper, the size of a hidden layer is 100, the word embedding dimensionality is 100 and the context word window size is 5.

We initialized all the square weight matrices as random orthogonal matrices. All the bias vectors were initialized to zero. Other weight matrices were sampled from a Gaussian distribution with mean 0 and variance $0.01^2$.

To update the model parameters, we have used Truncated Back-Propagation-Through-Time (T-BPTT)~\cite{Werbos:1990} (Werbos, 1990) with stochastic gradient descent. We fixed the depth of BPTT to 7 for all the models.  We used Ada-delta (Zeiler, 2012)~\cite{Zeiler:2002} to  adapt the learning rate of each parameter automatically ($\epsilon = 10^{-6}$ and $\rho = 0.95$). We trained each model for 50 epochs.

\section{Experiments and Results}
\label{sec:results}
In this section, we describe the experiments carried out for the shared task and present the experimental results.

\subsection{Data distribution}
We have used the corpus shared by the organizers for our experiments. The split for training/development/testing is detailed in Table~\ref{tab:data}. Test1 split was used for evaluating the different experiments that we have carried out for the shared task. Evaluation scores displayed in the results section are against Test1 only. Organizers provided another set of test data (Test2), against which all the submitted systems were evaluated.

\begin{table}[h] \small
\begin{center}
\begin{tabular}{|l|r|r|} 
\hline \bf & \bf \#sentences & \bf \#tokens \\ \hline
train & 11000 &  184697 \\
dev & 1000 & 17777 \\
test1 & 1000 & 16543 \\
test2 & 2000 &  34477\\ \hline
\end{tabular}
\end{center}
\caption{Corpus distribusion.}
\label{tab:data}
\end{table}

\subsection{Methodology}
In the following subsections, we discuss our approaches for word/phrase level quality estimation.

\subsubsection{Word Level QE}
Our experiments are mainly focused on the word level QE. We have used the output of the word level QE system for the estimation of the phrase level quality.

As mentioned above, we have used the modified RNN-LM architecture for the experiments. Baseline (LSTM) system was developed by training word embedding from scratch with other parameters of the model. In another set of experiments, we have pre-trained the word embedding with $word2vec$~\cite{Mikolov:MT1:2013}, and further tuned with the training of the model parameters. For pretraining, we have used an additional corpus (2M sentences approx.) from English-German Europarl data~\cite{koehn:2005}.

For bilingual models, we restructured the source sentence (English) according to the target (German) using word alignment provided by the organizers. For many-to-one mapping in the alignment (English-German), we chose the first alignment only. The `NULL' token was assigned to the words where were not aligned with any word on the target side. The input of the model is constructed by concatenating context words of source and target. For example, consider the source word sequence {$s_1s_2s_3$}, and the target word sequence {$t_1t_2t_3$}, then the input to the network will be {$s_1s_2s_3t_1t_2t_3$}.

In the training data, the distribution of the labels (OK/BAD) is skewed (OK to BAD ratio is approx. 4:1). To handle the issue, we tried one of the strategies proposed by~\newcite{shang:2015}, in which we replace `OK' label with sub-labels to balance the distribution. The sub-labels are OK\_B, OK\_I, OK\_E, depending on the location of the token in the sentence.

\subsubsection{Phrase Level QE}
For phrase level QE, we have not trained any explicit system. As it was mentioned by the organizers that a phrase is tagged as `BAD', if any word in the phrase is an incorrect translation. So, We have taken the output of the word level QE system and tagged the phrase as `BAD', if any word in the phrase boundary is tagged `BAD'. And other phrases (all words have the OK tag) are simply tagged as `OK'.

\begin{table}[h] \small
\begin{center}
\begin{tabular}{|l|r|r|}
\hline \bf Model/Test & \bf F1 BAD & \bf F1 OK \\ \hline
Baseline (LSTM) & 35.60 & 82.93 \\
LSTM\_PT & 37.27 & 83.25 \\ 
LSTM\_PT\_SL & 36.27 & 81.38 \\ \hline
LSTM\_BL & 36.18 & 82.51 \\ 
LSTM\_BL\_PT & 38.53 & 83.80 \\ 
LSTM\_BL\_PT\_SL & 39.17 & 83.20 \\ \hline
DeepLSTM & 35.86 & 80.35 \\ 
DeepLSTM\_PT & 36.81 & 82.51 \\
DeepLSTM\_PT\_SL & 36.13 & 81.32 \\ \hline
DeepLSTM\_BL & 37.41 & 81.92 \\ 
DeepLSTM\_BL\_PT & 38.38 & 81.41\\ 
DeepLSTM\_BL\_PT\_SL & 37.04 & 82.40 \\ \hline
GRU & 37.98 &  84.29 \\
GRU\_PT & 39.42  & 84.81 \\
GRU\_PT\_SL & \bf 40.46 & \bf 83.09 \\ \hline
GRU\_BL & 41.56 & 84.57 \\
GRU\_BL\_PT & 42.46 & 83.76 \\ 
GRU\_BL\_PT\_SL & 42.92 & 83.62 \\
\hline
\end{tabular}
\end{center}
\caption{F1 scores of different experiments for Word level QE. (PT: Pretrain; BL: Bilingual; SL: Sublabels).}
\label{tab:word-results} 
\end{table} 

\begin{table}[h] \small
\begin{center}
\begin{tabular}{|l|r|r|}
\hline \bf Model/Test & \bf F1 BAD & \bf F1 OK \\ \hline
Baseline (LSTM) & 43.46 & 75.41 \\
LSTM\_PT & 45.41 & 75.67 \\
LSTM\_PT\_SL & 44.92 & 73.11 \\ \hline
LSTM\_BL & 44.43 & 74.93 \\ 
LSTM\_BL\_PT & 45.75 & 77.17 \\ 
LSTM\_BL\_PT\_SL & 46.96 & 75.73 \\ \hline
DeepLSTM & 43.83 & 71.98 \\
DeepLSTM\_PT & 44.92 & 74.17 \\
DeepLSTM\_PT\_SL & 43.85 & 72.32 \\ \hline
DeepLSTM\_BL & 45.65 & 73.81 \\ 
DeepLSTM\_BL\_PT & 46.50 & 72.68 \\  
DeepLSTM\_BL\_PT\_SL & 45.63 & 74.57 \\ \hline
GRU & 45.70 & 77.86 \\
GRU\_PT & 46.49 & 80.00 \\
GRU\_PT\_SL & \bf 48.38 & \bf 76.14 \\ \hline
GRU\_BL & 48.11 & 77.69 \\
GRU\_BL\_PT & 49.58 & 76.88 \\ 
GRU\_BL\_PT\_SL & 49.61 & 77.20 \\ \hline
\end{tabular}
\end{center}
\caption{F1 scores of different experiments for Phrase level QE.}
\label{tab:phrase-results} 
\end{table} 

\subsection{Results}
To develop a baseline system for word and phrase level QE, organizers have used the baseline features (22 features) to train a Conditional Random Field (CRF) model with CRFSuite tool. The results of the experiments against Test2 are displayed in Table~\ref{tab:word-results} and~\ref{tab:phrase-results}.

We have evaluated our systems using the F1-score. As `OK' class is dominant in the data and a naive system tagging all the words `OK' will score high. Hence, F1-score of the `BAD' class has been used as a primary metric for the system evaluation. We have used the separate set of test and development corpus as shown in Table~\ref{tab:word-results}. The evaluation of all the experiments against Test1 corpus is displayed in Table~\ref{tab:word-results} for word level QE. Results for phrase level QE are shown in Table~\ref{tab:phrase-results}.

From the result tables, it is evident that GRU outperforms LSTM as reported by~\newcite{Cho:2014} for this task as well.
Pre-training is helpful in all the models. Also, the introduction of sub-labels is able to handle the problem of label-imbalance up to some extent. The results of Bilingual models are better than monolingual models, as reported by~\newcite{kreutzer:2015}.

\subsection{Submission to the shared task}
We have participated in the Task-2, which includes word and phrase level quality estimation. The submitted system setting was: $GRU+Pretrain+Sublabels$, which is \textbf{marked} in the result tables (2 and 3) as well. Table~\ref{tab:word-submission} and~\ref{tab:phrase-submission} detail the \footnotemark results \footnotetext{\url{http://www.quest.dcs.shef.ac.uk/wmt16_files_qe/wmt16_task2_results.pdf}} of the submission on Test2 corpus. The submission results were provided by the organizers.

\begin{table}[h] \small
\begin{center}
\begin{tabular}{|l|r|r|}
\hline \bf & \bf F1 BAD & \bf F1 OK \\ \hline
Baseline (CRF) & 36.82 & 88.00 \\
Submitted system & 41.92 & 84.21 \\
\hline
\end{tabular}
\end{center}
\caption{Results, word level submission.}
\label{tab:word-submission} 
\end{table} 

\begin{table}[h] \small
\begin{center}
\begin{tabular}{|l|r|r|}
\hline \bf & \bf F1 BAD & \bf F1 OK \\ \hline
Baseline (CRF) & 40.14 & 80.01 \\
Submitted system & 50.31 & 75.50 \\
\hline
\end{tabular}
\end{center}
\caption{Results, phrase level submission.}
\label{tab:phrase-submission} 
\end{table} 


\section{Discussion}
\label{sec:discussion}
The approach is language independent and it uses only context words' vector for predicting the tag for a word. In the other words, we check if any word fits (grammatically) in the given slot of words or not. We could use language specific features to enhance the classification accuracy, though. Experiments with bilingual models are similar to the concept of adding more features to any machine learning algorithm. In monolingual models, we use only target (German) words' vector as feature whereas, in bilingual models, we use source (English) words' vector also. A challenge which machine learning practitioners often face is, how to deal with skewed classes in classification problems. The distribution of classes (OK/BAD) is skewed in our case as well. To handle the issue, we tried to balance the distribution of classes by introducing the sub-labels.

LSTM and GRU are quite similar models, except the gating mechanism. It is hard to say which model will perform better in what conditions or in general~\cite{Chung:2014}. In this paper and in general as well, this restricts us to conduct only the empirical comparison between the LSTM and the GRU units. Deep models generally perform better than the shallow models, which is opposite for this task where LSTM outperforms Deep LSTM. The reason could be the insufficient data for training the deep models.

\section{Conclusion and Future Work}
\label{sec:conc}
We have developed a language independent word/phrase level Quality Estimation system using RNN. We have used RNN-LM architecture, with LSTM, deep LSTM, and GRU. We showed that these models benefit from pretraining and the introduction of sub-labels. Also, models with bilingual features outperform the monolingual models.

We can extend the work for sentence and document level quality estimation. Improving the word level quality estimation with data selection and bootstrapping (Logacheva et al., 2015), more effective ways to handle label-imbalance, training bigger models, using language specific features, other variations of LSTM architecture etc., are the other possibilities. 

\bibliographystyle{acl2016}
\bibliography{wmt2016}

\end{document}